\documentclass[12pt,reqno]{amsart}
\usepackage{fullpage}
\usepackage{amsmath}
\usepackage{mathrsfs,amssymb,graphicx,verbatim}
\usepackage{paralist}
\usepackage{amsfonts}
\usepackage{amssymb}
\usepackage{multirow}
\usepackage{euler, latexsym}

\usepackage{ifthen}

\newcommand{\ifsodaelse}[2]{\ifthenelse{\isundefined{\SODAF}}{#2}{#1}}

\ifsodaelse    {\usepackage{ltexpprt}\usepackage{amsmath}}{}
\usepackage{mathrsfs,amssymb,graphicx,verbatim}
\usepackage{paralist}
\usepackage[breaklinks,colorlinks,pdfstartview=FitH]{hyperref}

\newcommand\remove[1]{}

\newcommand{\rnote}[1]{}
\newcommand{\jnote}[1]{}

\date{}

\include{psfig}

\title[MICCAI-2011 Paper]{Learning Shape and Texture Characteristics of CT Tree-in-Bud Opacities  for CAD Systems}

\author{Ula\c{s}~Ba\u{g}c{\i},  Jianhua Yao, Jesus Caban,  Anthony F. Suffredini, Tara N. Palmore, Daniel J. Mollura}
\address{Radiology and Imaging Sciences\\ National Institutes of Health}
\email{ulasbagc@gmail.com}

\date{}

\begin{document}
\maketitle

\begin{abstract} 
Although radiologists can employ CAD systems to characterize malignancies, pulmonary fibrosis and other chronic diseases; the design of imaging techniques to quantify infectious diseases continue to lag behind. There exists a need to create more CAD systems capable of detecting and quantifying characteristic patterns often seen in respiratory tract infections such as influenza, bacterial pneumonia, or tuborculosis. One of such patterns is Tree-in-bud (TIB) which presents \textit{thickened} bronchial structures surrounding by clusters of \textit{micro-nodules}. Automatic detection of TIB patterns is a challenging task because of their weak boundary, noisy appearance, and small lesion size. In this paper, we present two novel methods for automatically detecting TIB patterns: (1) a fast localization of candidate patterns using information from local scale of the images, and (2) a M\"{o}bius invariant feature extraction method based on learned local shape and texture properties. A comparative evaluation of the proposed methods is presented with a dataset of 39 laboratory confirmed viral bronchiolitis human parainfluenza (HPIV) CTs and 21 normal lung CTs. Experimental results demonstrate that the proposed CAD system can achieve high detection rate with an overall accuracy of 90.96\%.  
\end{abstract}

\setcounter{tocdepth}{2}
\tableofcontents

\section{Introduction}
\begin{figure}
\begin{center}
\includegraphics[height=5.1 cm]{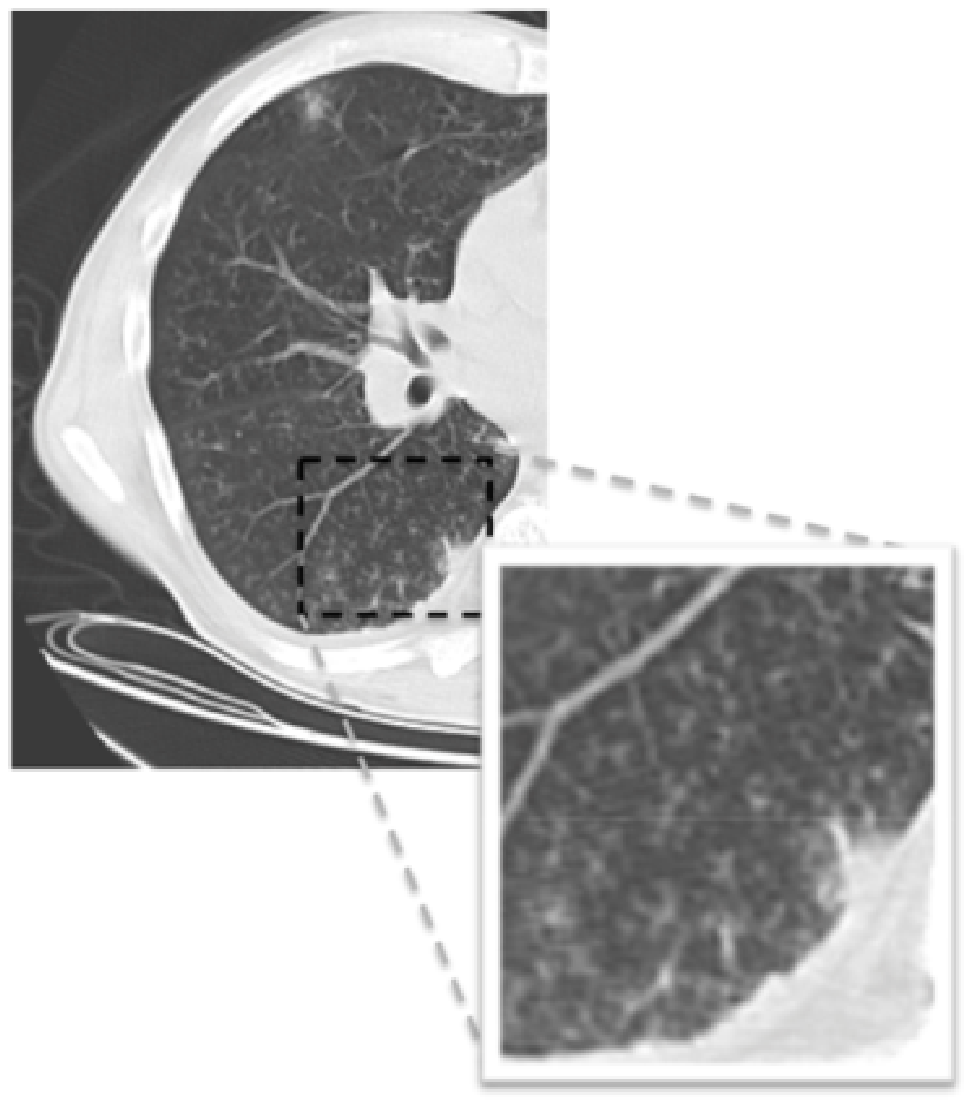}
\includegraphics[height=5.0 cm]{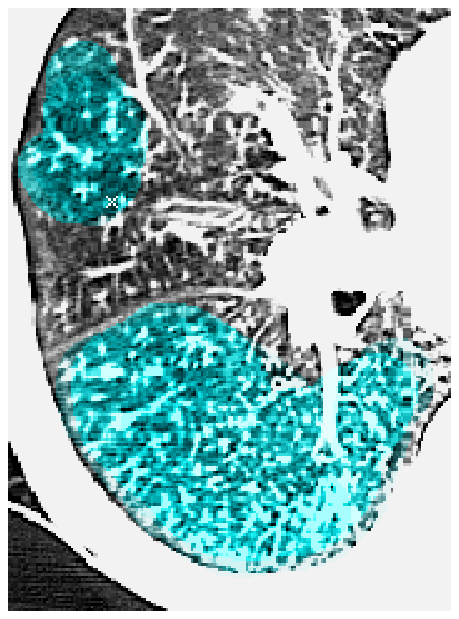}
\end{center} 
\caption{(Left) CT image with a significant amount TIB patterns. (Right) Labelled TIB patterns (blue) in zoomed window on the right lung.\label{img:tib}}
\end{figure}

As shown by the recent pandemic of novel swine-origin H1N1 influenza, respiratory tract infections are a leading cause of disability and death. A common image pattern often associated with respiratory tract infections is  TIB opacification, represented by thickened bronchial structures locally surrounded by clusters of 2-3 millimeter micro-nodules. Such patterns generally represent disease of the small airways such as infectious-inflammatory bronchiolitis as well as bronchiolar luminal impaction with mucus, pus, cells or fluid causing normally invisible peripheral airways to become visible~\cite{eisenhuber}. Fig.~\ref{img:tib} shows TIB patterns in a chest CT.

The precise quantification of the lung volume occupied by TIB patterns is a challenging task limited by significant inter-observer variance with inconsistent visual scoring methods. These limitations raise the possibility that radiologist{s'} assessment of respiratory tract infections could be enhanced through the use of computer assisted detection (CAD) systems. However, there are many technical obstacles to detecting TIB patterns because micro-nodules and abnormal peripheral airway structures have strong shape and appearance similarities to TIB patterns and normal anatomic structures in the lungs.  

In this work, we propose a new CAD system to evaluate and quantify respiratory tract infections by automatically detecting TIB patterns. The main contributions of the paper are two-fold: (1) A candidate selection method that locates possible abnormal patterns in the images. This process comes from a learning perspective such that the size, shape, and textural characteristics of TIB patterns are learned a priori. The candidate selection process removes large homogeneous regions from consideration which results in a fast localization of candidate TIB patterns. The local regions enclosing candidate TIB patterns are then used to extract shape and texture features for automatic detection; (2) another novel aspect in this work is to extract M\"{o}bius invariant local shape features. Extracted local shape features are combined with statistical texture features to classify lung tissues. To the best of our knowledge, this is the first study that uses automatic detection of TIB patterns for a CAD system in infectious lung diseases. Since there is no published work on automatic detection of TIB patterns in the literature, we compare our proposed CAD system on the basis of different feature sets previously shown to be successful in detecting lung diseases in general.

\section{Methodology}
\label{sec:prep}
\begin{figure}
  \begin{center}
   \includegraphics[height=7.0 cm]{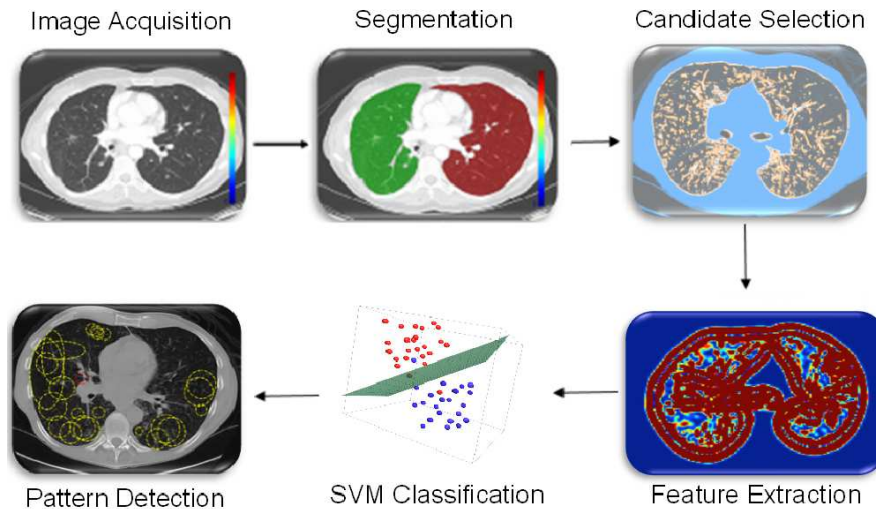}
   \end{center}
\caption{The flowchart of the proposed CAD system for automatic TIB detection. \label{img:overall}}
\end{figure}

The proposed CAD methodology is illustrated in Fig.~\ref{img:overall}. First, lungs are segmented from CT volumes. Second,  we use locally adaptive scale based filtering method to detect candidate TIB patterns. Third, segmented lung is divided into local patches in which we extract invariant shape features and statistical texture features followed by support vector machine (SVM) classification. We extract features from local patches of the segmented lung only if there are candidate TIB patterns in the patches. The details of the proposed methods are presented below.\\

\noindent \textbf{I. Segmentation:} Segmentation is often the first step in CAD systems. There are many clinically accepted segmentation methods in clinics~\cite{sfc,hu}. In this study, fuzzy connectedness (FC) image segmentation algorithm is used to achieve successful delineation~\cite{sfc}. In FC framework, left and right lungs are ``recognized" by automatically assigned seeds, which initiate FC segmentation.\\ 

\noindent \textbf{II. Learning characteristics of TIB patterns:} From Fig.~\ref{img:tib}, we can readily observe that TIB patterns have intensity characteristics with high variation towards nearby pixels, and such regions do not usually exceed a few millimetre(mm) in length. In other words, TIB patterns do not constitute sufficiently large homogeneous regions. Non-smooth changes in local gradient values support this observation. As guided by these observations, we conclude that (a) TIB patterns are \textit{localized only in the vicinity of small homogeneous regions}, and (b) their boundaries have high curvatures due to the nature of its complex shape.\\

\noindent \textbf{III. Candidate Pattern Selection:}  Our candidate detection method comes from a learning perspective such that we assign every internal voxel of the lung a membership value reflecting the size (i.e., scale) of the homogeneous region that the voxel belongs to. To do this, we use a locally adaptive scale based filtering method called ball-scale (or b-scale for short)~\cite{sfc}. b-scale is the \textit{simplest} form of a locally adaptive scale where the scene is partitioned into  several scale levels within which every voxel is assigned the size of the local structure it belongs. For instance, voxels within the large homogeneous objects have highest scale values, and the voxels nearby the boundary of objects have small scale values. Because of this fact and the fact in II.(a), we draw the conclusion that TIB patterns constitute only small b-scale values, hence, it is highly reasonable to consider voxels with small b-scale values as candidate TIB patterns. Moreover, it is indeed highly practical to discard voxels with high b-scale values from candidate selection procedure. Fig.~\ref{img:overall} (candidate selection) and Fig.~\ref{img:shapes}(b) show selected b-scale regions as candidate TIB patterns. A detailed description of the b-scale algorithm is presented in~\cite{sfc}. 

\section{Feature Extraction}
For a successful CAD system for infectious lung diseases, there is a need to have representative features characterizing shape and texture of TIB patterns efficiently. Since TIB is a complex shape pattern consisting of curvilinear structures with nodular structures nearby (i.e., a budding tree), we propose to use local shape features (derived from geometry of the local structures) combined with grey-level statistics (derived from a given local patch). 

It has been long known that curvatures play an important role in the representation and recognition of intrinsic shapes. However, similarity of curvature values may not necessarily be equivalent to intrinsic shape similarities, which causes a degradation in recognition and matching performance. To overcome this difficulty, we propose to use  Willmore energy functional~\cite{willmore} and several different affine invariant shape features parametrically related to the Willmore energy functional. 

\subsubsection{Willmore Energy:} The Willmore energy of surfaces plays an important role in digital geometry, elastic membranes, and image processing. It is closely related to Canham-Helfrich model, where surface energy is defined as
\begin{equation}
\mathcal{S}=\int_{\Sigma}\alpha+\beta (H)^2-\gamma KdA.
\end{equation}
This model is curvature driven, invariant under the the group of M\"{o}bius transformations (in particular under rigid motions and scaling of the surface) and shown to be very useful in energy minimization problems. Invariance of the energy under rigid motions leads to conservation of linear and angular momenta, and invariance under scaling plays a role in setting the size of complex parts of the intrinsic shapes (i.e., corners, wrinkles, folds). In other words, the position, grey-level characteristics, size and orientation of the pattern of interest have minimal effect on the extracted features as long as the suitable patch is reserved for the analysis. In order to have simpler and more intuitive representation of the given model, we simply set $\alpha=0$ and $\beta=\gamma=1$, and the equation turns into the Willmore energy functional, 
\begin{equation}
\mathcal{S}_w=\int_{\Sigma}(H^2-K)dA=\int_{\Sigma}|H|^2dA-\int_{\partial \Sigma}|K|ds,
\end{equation}
where $H$ is the mean curvature vector on $\Sigma$, $K$ the Gaussian curvature on $\partial \Sigma$, and $dA$, $ds$ the induced area and length metrics on $\Sigma$, $\partial \Sigma$ (representing area and boundary, respectively). Since homogeneity region that a typical TIB pattern appears is small in size, total curvature (or energy) of that region is high and can be used as a discriminative feature. 

In addition to Willmore energy features, we have included seven different local shape features in the proposed CAD system. Let $\kappa_1$ and $\kappa_2$ indicate eigenvalues of the local Hessian matrix for any given local patch, the following shape features are extracted: 1) mean curvature ($H$), 2) Gaussian curvature ($K$), 3) shape index ($SI$), 4) elongation ($\kappa_1/\kappa_2$), 5) shear ($(\kappa_1-\kappa_2)^2/4$), 6) compactness ($1/(\kappa_1\kappa_2)$), and 7) distortion ($\kappa_1-\kappa_2$).
 Briefly, the $SI$ is a statistical measure used to define local shape of the localized structure within the image~\cite{griffin}. 
 
\begin{figure}
  \begin{center}
   \includegraphics[height=6.0 cm]{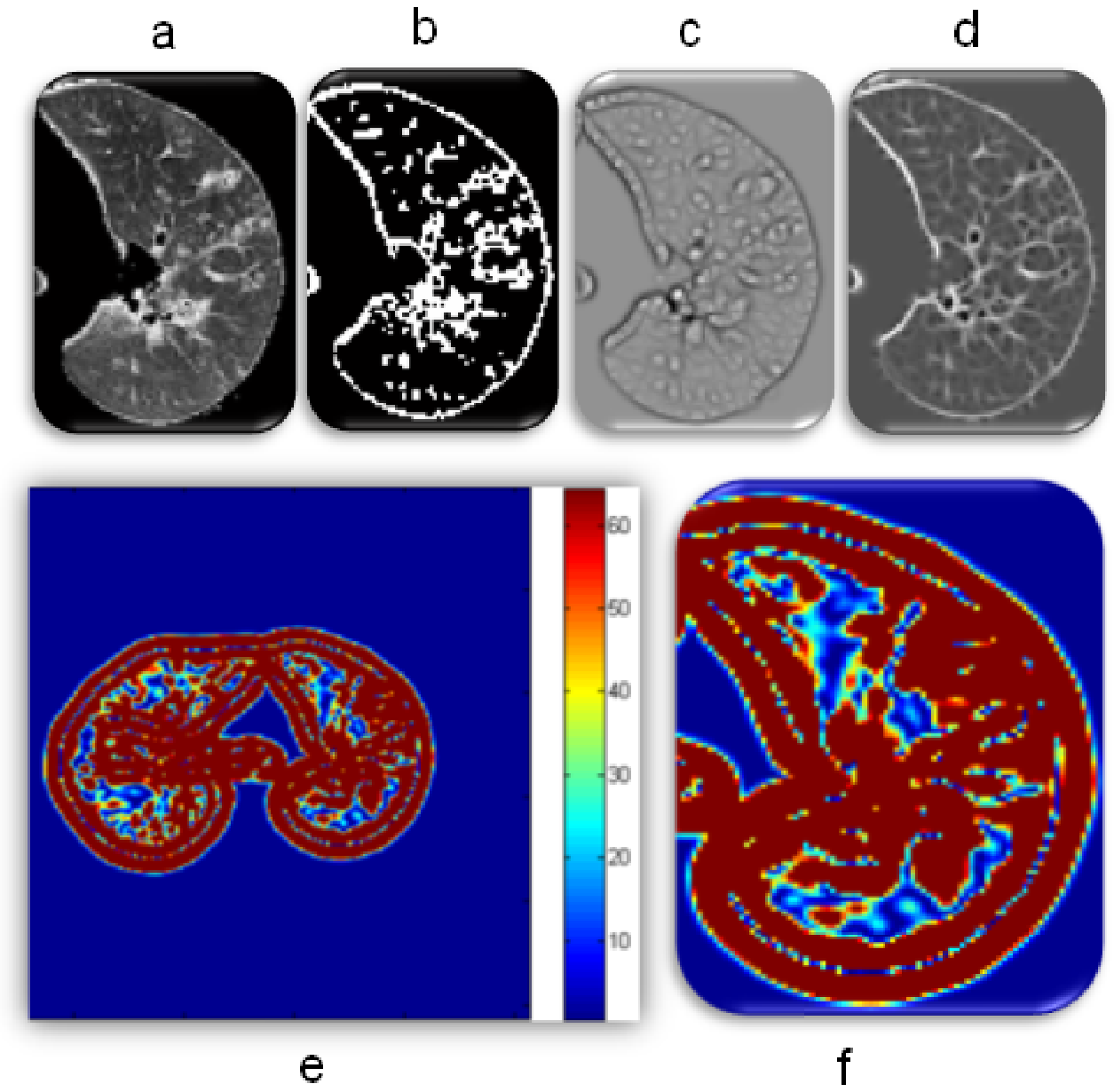} 
    \includegraphics[height=5.6 cm]{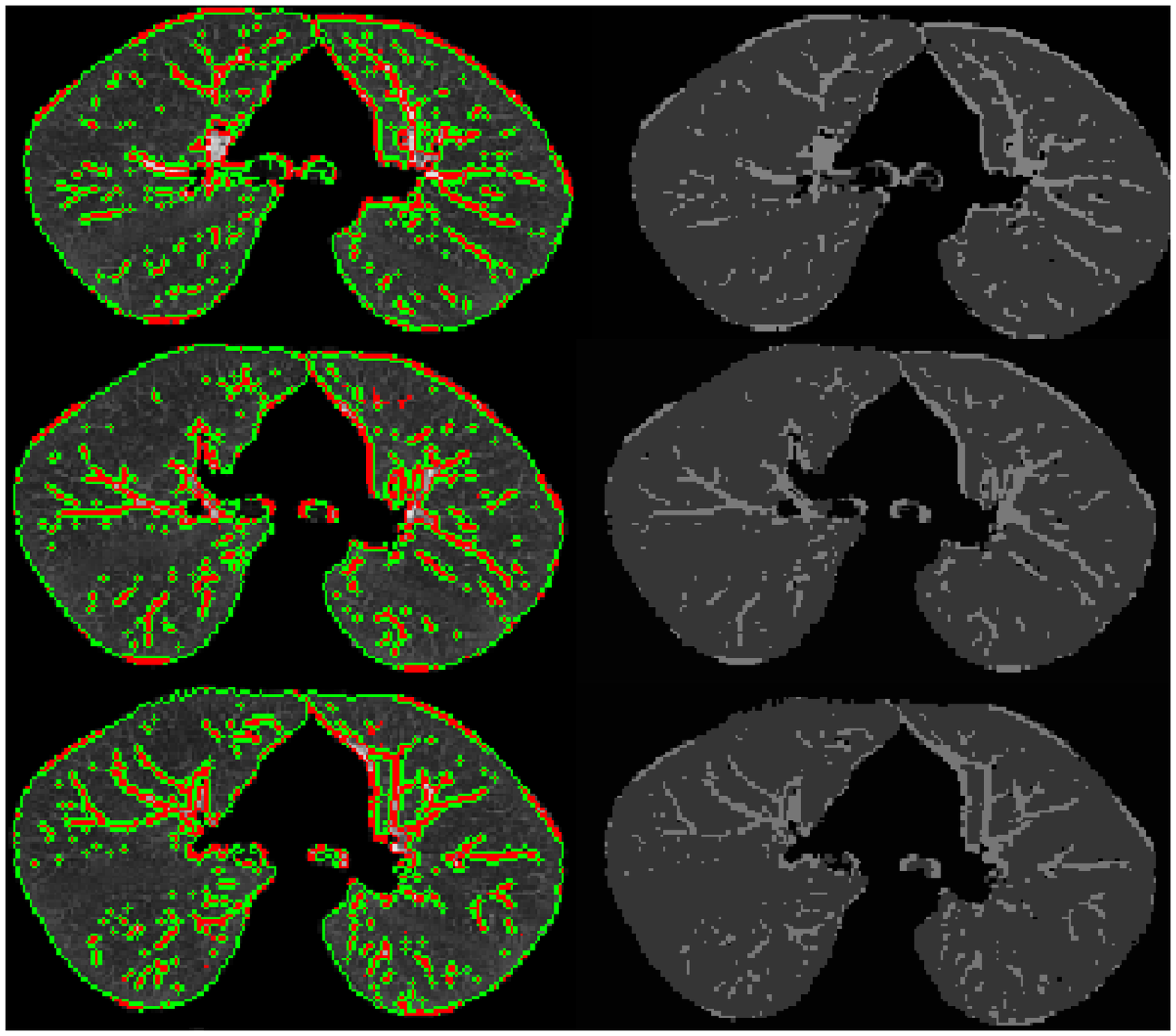}\\
    \quad  \quad  \quad  \quad  \quad  \quad  \quad  \quad  \quad  \quad  \quad  \quad  \quad  \quad  \quad  \quad  g
   \end{center}
\caption{a. CT lung, b. selected b-scale patterns, c. mean Curvature map ($H$), d. Gaussian Curvature ($K$), e. Willmore energy map, f. zoomed (e). g. Multi-phase level set segmentation based on the proposed shape features is shown in three different slices from the same patient's chest CT scan.\label{img:shapes}}
\end{figure}

Elongation indicates the \textit{flatness} of the shape. Compactness feature measures the similarity between shape of interest and a perfect ellipse. Fig.~\ref{img:shapes}(c) and (d) show mean and Gaussian curvature maps from which all the other local shape features are extracted. Fig.~\ref{img:shapes}(e) and (f) show  Willmore energy map extracted from Fig.~\ref{img:shapes}(a). 

Based on the observation in training, TIB patterns most likely occur in the regions inside the lung with certain ranges (i.e, blue and yellow regions). This observation facilitates one practically useful fact in the algorithm that, in the feature extraction process, \textit{we only extract features if and only if at least ``one" b-scale pattern exists in the local region as well as Willmore energy values of pixels lie in the interval observed from training.} Moreover, considering the Willmore energy has a role as hard control on feature selection and computation, it is natural to investigate their ability to segment images. We present a segmentation framework in which every voxel is described by the proposed shape features. A multi-phase level set~\cite{ayed} is then applied to the resulting vectorial image and the results are shown in Fig.~\ref{img:shapes}(g). First and second columns of the Fig.~\ref{img:shapes}(g) show segmented structures and the output homogeneity maps showing segmented regions in different grey-level, respectively. Although segmentation of small airway structures and pathological patterns is not the particular aim of this study, the proposed shape features show promising results due to their discriminative power.\\ 

\noindent {\bf Texture features:}  Spatial statistics based on Grey-Level Co-occurrence Matrix (GLCM)~\cite{glcm} are shown to be useful in discriminating patterns pertaining to lung diseases. As texture can give a lot of insights into the classification and characterization problem of poorly defined lesions, regions, and objects, we combine our proposed shape based invariants with GLCM based features. We extract 18 GLCM features from each local patch including autocorrelation,  entropy, variance, homogeneity, and extended features of those. Apart from the proposed method, we also compare our proposed method with well known texture features: steerable wavelets (computed over 1 scale and 6 orientations with derivative of Gaussian kernel), GLCM, combination of shape and steerable wavelets, and considering different local patch size.

 \section{Experimental Results}
39 laboratory confirmed CTs of HPIV infection and 21 normal lung CTs were collected for the experiments. The in-plane resolution is affected from patients' size and varying from 0.62mm to 0.82mm with slice thickness of 5mm. An expert radiologist carefully examined the complete scan and labeled the regions as normal and abnormal (with TIB patterns). As many regions as possible showing abnormal lung tissue were labeled (see Table~\ref{table:roctable} for details of the number of regions used in the experiments). After the proposed CAD system is tested via two-fold cross validations with labeled dataset, we present receiver operator characteristic (ROC) curves of the system performances.

\begin{figure}
\begin{center}
\includegraphics[height =7.5 cm]{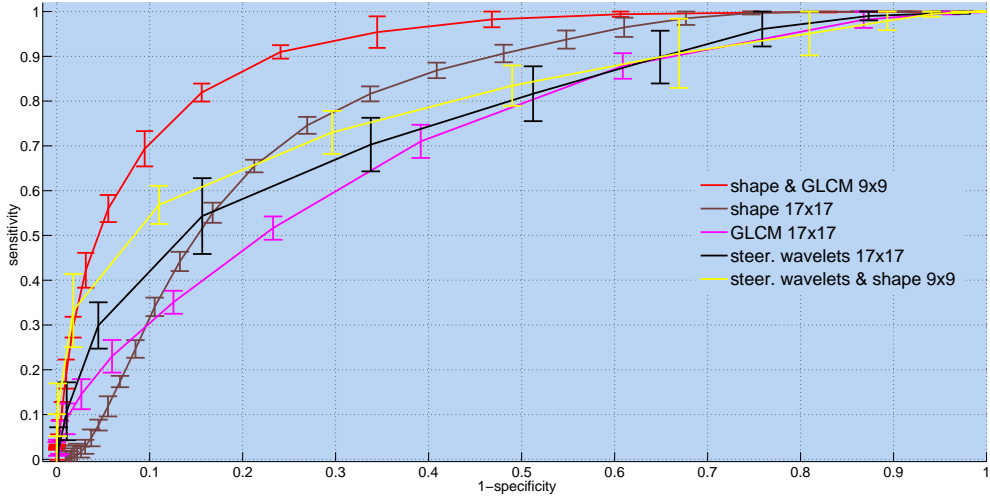}
\caption{Comparison of CAD performances via ROC curves of different feature sets.\label{img:roc1}}
\end{center}
\end{figure}

\begin{table}
\caption{Accuracy ($A_z$) of the CAD system with given feature sets.\label{table:roctable}}
\begin{center}
\begin{tabular}{|c|c|c|c|c|c|}
\hline \begin{scriptsize}\textbf{Features} \end{scriptsize}& \begin{scriptsize}
\textbf{Dimension}\end{scriptsize} & \begin{scriptsize}\textbf{Patch Size}\end{scriptsize}&\begin{scriptsize}\textbf{\# of  patches}\end{scriptsize}&\begin{scriptsize}\textbf{\# of  patches}\end{scriptsize}& \begin{scriptsize}
\textbf{Area under} 
\end{scriptsize}\\ 
& & & \begin{scriptsize}
\textbf{(TIB)}
\end{scriptsize}  &\begin{scriptsize}
\textbf{(Normal)}
\end{scriptsize}  &\begin{scriptsize}
\textbf{ROC curve: $A_z$}
\end{scriptsize}\\ \hline \hline
\begin{scriptsize}Shape \& GLCM\end{scriptsize} & \begin{scriptsize}
8+18=26\end{scriptsize} & \begin{scriptsize}
17x17\end{scriptsize} & \begin{scriptsize}
14144
\end{scriptsize} & \begin{scriptsize}
12032
\end{scriptsize} &\begin{scriptsize}
\textbf{0.8991}\end{scriptsize}\\ 
\begin{scriptsize}
Shape \& GLCM\end{scriptsize} & \begin{scriptsize}
8+18=26\end{scriptsize} & \begin{scriptsize}
13x13\end{scriptsize} & \begin{scriptsize}
24184 
\end{scriptsize}& \begin{scriptsize}
20572 
\end{scriptsize}&\begin{scriptsize}
\textbf{0.9038}\end{scriptsize}\\ 
\begin{scriptsize}
Shape \& GLCM \end{scriptsize}& \begin{scriptsize}
8+18=26\end{scriptsize} & \begin{scriptsize}9x9 
\end{scriptsize}& \begin{scriptsize}
50456 
\end{scriptsize}& \begin{scriptsize}
42924
\end{scriptsize} &\begin{scriptsize}
\textbf{0.9096}\end{scriptsize}\\ \hline
\begin{scriptsize}
Shape \end{scriptsize} & \begin{scriptsize}
8\end{scriptsize}  & \begin{scriptsize}
17x17\end{scriptsize} & \begin{scriptsize}
14144
\end{scriptsize} &\begin{scriptsize}
 12032
\end{scriptsize} &\begin{scriptsize}
0.7941\end{scriptsize}\\ 
\begin{scriptsize}
Shape\end{scriptsize}  & \begin{scriptsize}
8\end{scriptsize}  & \begin{scriptsize}
13x13
\end{scriptsize} &\begin{scriptsize}
 24184
\end{scriptsize} &\begin{scriptsize}
 20572 
\end{scriptsize}& \begin{scriptsize}
0.7742\end{scriptsize}\\ 
\begin{scriptsize}
Shape \end{scriptsize} & \begin{scriptsize}
8\end{scriptsize} & \begin{scriptsize}
9x9\end{scriptsize} & \begin{scriptsize}
 50456
\end{scriptsize} & \begin{scriptsize}
42924
\end{scriptsize} & \begin{scriptsize}
0.7450\end{scriptsize}\\ \hline
\begin{scriptsize}
Steer. Wavelets\& Shape 
\end{scriptsize} &\begin{scriptsize}
 6x17x17+8=1742
\end{scriptsize}  & \begin{scriptsize}
17x17
\end{scriptsize} & \begin{scriptsize}
14144
\end{scriptsize} &\begin{scriptsize}
 12032
\end{scriptsize} & \begin{scriptsize}
0.7846
\end{scriptsize}\\ 
\begin{scriptsize}
Steer. Wavelets\& Shape
\end{scriptsize}  & \begin{scriptsize}
6x13x13+8=1022
\end{scriptsize}  & \begin{scriptsize}
13x13 \end{scriptsize}& \begin{scriptsize}
24184
\end{scriptsize}& \begin{scriptsize}
20572
\end{scriptsize}& \begin{scriptsize}0.7692
\end{scriptsize}\\ 
\begin{scriptsize}
Steer. Wavelets\& Shape
\end{scriptsize}  & \begin{scriptsize}
6x9x9+8=494
\end{scriptsize} & \begin{scriptsize}
9x9
\end{scriptsize} &\begin{scriptsize}
50456
\end{scriptsize} & \begin{scriptsize}
42924 
\end{scriptsize}& \begin{scriptsize}
0.7908
\end{scriptsize}\\ \hline
\begin{scriptsize}
Steer. Wavelets
\end{scriptsize}  & \begin{scriptsize}
6x17x17=1734
\end{scriptsize}  & \begin{scriptsize}
17x17
\end{scriptsize} & \begin{scriptsize}
14144
\end{scriptsize} & \begin{scriptsize}
12032
\end{scriptsize}& \begin{scriptsize}
0.7571
\end{scriptsize}\\ 
\begin{scriptsize}
Steer. Wavelets 
\end{scriptsize} & \begin{scriptsize}
6x13x13=1014 
\end{scriptsize} & \begin{scriptsize}
13x13
\end{scriptsize} & \begin{scriptsize}
24184
\end{scriptsize} & \begin{scriptsize}
20572
\end{scriptsize} & \begin{scriptsize}
0.7298
\end{scriptsize}\\ 
\begin{scriptsize}
Steer. Wavelets 
\end{scriptsize} & \begin{scriptsize}
6x9x9=486
\end{scriptsize} & \begin{scriptsize}
9x9 
\end{scriptsize}& \begin{scriptsize}
50456
\end{scriptsize} &  \begin{scriptsize}
42924
\end{scriptsize} & \begin{scriptsize}
0.7410
\end{scriptsize}\\ \hline
\begin{scriptsize}
GLCM
\end{scriptsize} & \begin{scriptsize}
18 
\end{scriptsize}& \begin{scriptsize}
17x17
\end{scriptsize} & \begin{scriptsize}
14144
\end{scriptsize} & \begin{scriptsize}
12032
\end{scriptsize} & \begin{scriptsize}
0.7163
\end{scriptsize}\\ 
\begin{scriptsize}
GLCM
\end{scriptsize} & \begin{scriptsize}
18
\end{scriptsize} & \begin{scriptsize}
13x13 
\end{scriptsize}& \begin{scriptsize}
24184
\end{scriptsize} &\begin{scriptsize}
 20572
\end{scriptsize} & \begin{scriptsize}
0.7068
\end{scriptsize}\\ 
\begin{scriptsize}
GLCM 
\end{scriptsize}& \begin{scriptsize}
18 
\end{scriptsize}& \begin{scriptsize}
9x9 
\end{scriptsize}& \begin{scriptsize}
50456
\end{scriptsize} &\begin{scriptsize}
 42924
\end{scriptsize} & \begin{scriptsize}
0.6810
\end{scriptsize}\\ \hline
\end{tabular}
\end{center}
\end{table}

\begin{table}
\caption{p-values are shown in confusion matrix.\label{table:pval}}
\begin{center}
\begin{tabular}{|c|c|c|c|c|}
\hline \begin{scriptsize}
\textbf{p-Confusion}
\end{scriptsize} &\begin{scriptsize}
 \textbf{Shape}
\end{scriptsize} & \begin{scriptsize}
\textbf{Steer.\& } 
\end{scriptsize}&\begin{scriptsize}
 \textbf{Steer.} 
\end{scriptsize}& \begin{scriptsize}
\textbf{GLCM}
\end{scriptsize} \\ 
\begin{scriptsize}
\textbf{Matrix}
\end{scriptsize}& &\begin{scriptsize}
 \textbf{Shape}
\end{scriptsize} & &\\ \hline
\begin{scriptsize}
\textbf{Shape\&}
\end{scriptsize} &  &  &  &  \\
\begin{scriptsize}
\textbf{GLCM}
\end{scriptsize} &\begin{scriptsize}
0.0171 
\end{scriptsize}& \begin{scriptsize}
0.0053
\end{scriptsize} & \begin{scriptsize}
0.0056
\end{scriptsize} & \begin{scriptsize}
0.0191
\end{scriptsize}\\ \hline
\begin{scriptsize}
\textbf{Shape}
\end{scriptsize} &-- & \begin{scriptsize}
0.0086
\end{scriptsize} & \begin{scriptsize}
0.0094
\end{scriptsize} & \begin{scriptsize}
0.0185
\end{scriptsize}\\ \hline
\begin{scriptsize}
\textbf{Steer.\&}
\end{scriptsize} &-- &-- & \begin{scriptsize}
0.0096
\end{scriptsize} & \begin{scriptsize}
0.0175
\end{scriptsize} \\
\begin{scriptsize}
\textbf{Shape}
\end{scriptsize} & & & & \\ \hline
\begin{scriptsize}
\textbf{Steer.}
\end{scriptsize} &-- &-- &-- & \begin{scriptsize}
0.0195
\end{scriptsize}\\ \hline
\end{tabular}
\end{center}
\end{table}
Table~\ref{table:roctable} summarizes the performance of the proposed CAD system as compared to different feature sets. The performances are reported as the areas under the ROC curves $(A_z)$. Note that shape features alone are superior to other methods even though the dimension of the shape feature is only 8. The best performance is obtained when we combined shape and GLCM features. This is expected because  spatial statistics are incorporated into the shape features such that texture and shape features are often complementary to each other. In what follows, we select the best window size for each feature set and plot their ROC curves all in Fig.~\ref{img:roc1}. To have a valid comparison, we repeat candidate selection step for all the methods, hence, the CAD performances of compared feature sets might perhaps have lower accuracies if the candidate selection part is not applied. Superiority of the proposed features is clear in all cases. To show whether the proposed method is significantly different than the other methods, we compared the performances through paired t-tests, and the p-values of the tests are summarized in Table~\ref{table:pval}. Note that statistically significant changes are emphasized by $p<.01$ and $p<.05$.

\section{Conclusion}
In this paper, we have proposed a novel CAD system for automatic TIB pattern detection from lung CTs. The proposed system integrates 1) fast localization of candidate TIB patterns through b-scale filtering and scale selection, and 2) combined shape and textural features to identify TIB patterns. Our proposed shape features illustrate the usefulness of the invariant features, Willmore energy features in particular, to analyze TIB patterns in Chest CT. In this paper, we have not addressed the issue of quantitative evaluation of severity of diseases by expert observers. This is a challenging task for complex shape patterns such as TIB opacities, and subject to further investigation.

\end{document}